\pdfoutput=1

\documentclass[11pt]{article}

\usepackage{acl}

\usepackage{times}
\usepackage{latexsym}

\usepackage[T1]{fontenc}

\usepackage[utf8]{inputenc}

\usepackage{microtype}

\usepackage{inconsolata}

\usepackage{graphicx}

\usepackage{subfig}
\usepackage{amsmath,amssymb,amsfonts}
\usepackage{booktabs}
\usepackage{tabularx}
\usepackage{multicol}
\usepackage{array}
\usepackage{seqsplit}
\usepackage{tcolorbox}
\usepackage{seqsplit}
\usepackage{float}
\usepackage{fixltx2e}
\usepackage{pgffor} 

\title{SPARKLE: Enhancing SPARQL Generation with \\ Direct KG Integration in Decoding}

\author{Jaebok Lee \and Hyeonjeong Shin \\
        Samsung Research, Seoul, South Korea \\
        \texttt{\{jaebok44.lee, hjjeong.shin\}@samsung.com} \\
        }

\begin{document}
\maketitle
\begin{abstract}
Existing KBQA methods have traditionally relied on multi-stage methodologies, involving tasks such as entity linking, subgraph retrieval and query structure generation.
However, multi-stage approaches are dependent on the accuracy of preceding steps, leading to cascading errors and increased inference time.
Although a few studies have explored the use of end-to-end models, they often suffer from lower accuracy and generate inoperative query that is not supported by the underlying data.
Furthermore, most prior approaches are limited to the static training data, potentially overlooking the evolving nature of knowledge bases over time.
To address these challenges, we present a novel end-to-end natural language to SPARQL framework, SPARKLE.
Notably SPARKLE leverages the structure of knowledge base directly during the decoding, effectively integrating knowledge into the query generation.
Our study reveals that simply referencing knowledge base during inference significantly reduces the occurrence of inexecutable query generations.
SPARKLE achieves new state-of-the-art results on SimpleQuestions-Wiki and highest F1 score on LCQuAD 1.0 (among models not using gold entities), while getting slightly lower result on the WebQSP dataset.\footnote{https://github.com/zzaebok/sparkle}
Finally, we demonstrate SPARKLE's fast inference speed and its ability to adapt when the knowledge base differs between the training and inference stages.
\end{abstract}

\section{Introduction}\label{introduction}

Knowledge Base Question Answering (KBQA) is a task which aims to answer user queries by extracting relevant information from structured knowledge bases, such as DBPedia \cite{dbpedia}, Freebase \cite{freebase} and Wikidata \cite{wikidata}.
KBQA systems enable users to interact with abundant information in knowledge bases without requiring an in-depth understanding of query languages.
Existing KBQA approaches have often adopted multi-stage pipelines, which involve entity linking frameworks such as DPR \cite{karpukhin-etal-2020-dense}, BLINK \cite{wu-etal-2020-scalable}, and ELQ \cite{li-etal-2020-efficient} or subgraph retrieval mechanisms like KNN and Personalized PageRank \cite{ppr} to handle the complexity of large-scale data sources.
While these strategies have shown promise in improving accuracy, they introduce undesirable side effects, especially an increase in inference time and a dependency on the performance of preceding modules \cite{yu2023decaf}.

In contrast to multi-stage methods, some studies have explored alternative approaches to KBQA.
For instance, end-to-end pre-trained language models (PLMs) have been employed to generate query language \cite{yin2021neural,sgpt} or direct answers among entities \cite{saffari-etal-2021-end,McKenna2023}.
While end-to-end semantic parsing models offer simplicity, they may produce invalid queries, as they lack an understanding of valid facts and connections in the knowledge base.
Moreover, these models are often tightly bound to their training data, struggling to adapt to the evolving nature of knowledge.

In this paper, we present a novel solution to address the aforementioned challenges in KBQA.
We propose a straightforward yet effective end-to-end natural language to SPARQL model for KBQA.
Building upon the concept of entity and relation retrieval within a generation model \cite{genre,GenRL}, we extend this approach to SPARQL query generation.
Our model not only generates entities and relations within a SPARQL query, but also directly leverages the structural information in the knowledge base to generate valid triple patterns.
This is achieved through the constrained decoding within a single sequence-to-sequence model.
Such an integration is natural and seamless, allowing left-to-right decoding process to accurately reflect the semantic structure of the knowledge base.

We empirically evaluate SPARKLE on three benchmark datasets: WebQSP \cite{webqsp}, SimpleQuestions-Wiki \cite{simplequestions}, and LCQuAD 1.0 \cite{lcquad1}.
Our model achieves new state-of-the-art result on SimpleQuestions-Wiki (+3.5 F1 score), and competitive result on LCQuAD 1.0 (the highest F1 score among models that do not use gold entities).
While its performance on WebQSP slightly lags behind, it still achieves the highest Hits@1 score among end-to-end methods, a point we will discuss in Section \ref{sec:overall}.
SPARKLE not only delivers fast inference times (under 1 second), suitable for real-world scenarios, but also supports batch processing of multiple questions simultaneously.
Additionally, we demonstrate that SPARKLE enables seamless adaptation to updated knowledge bases during inference without retraining.
Our model, initially trained with 2016-04 DBPedia, successfully retrieves newly added facts about events occurring between April and October 2016, by simply switching the knowledge base for inference to the 2016-10 DBPedia.
The contributions of this paper are summarized as follows:

\begin{itemize}
    \item We propose a novel end-to-end method for natural language to SPARQL translation through the constrained decoding in a single sequence-to-sequence model. To the best of our knowledge, no existing method uses the structure of knowledge base directly within decoding.
    \item SPARKLE is evaluated on three distinct KBQA datasets, each linked to a different knowledge base. Experimental results indicate that our approach achieves new state-of-the-art or competitive performance on SimpleQuestions-Wiki and LCQuAD 1.0, while showing fast inference speed and enabling batch processing.
    \item Further analysis demonstrates our model's capability to adapt to evolving knowledge base during inference without additional training.
\end{itemize}

\begin{figure*}[t]
    \centering
    \includegraphics[width=0.8\linewidth]{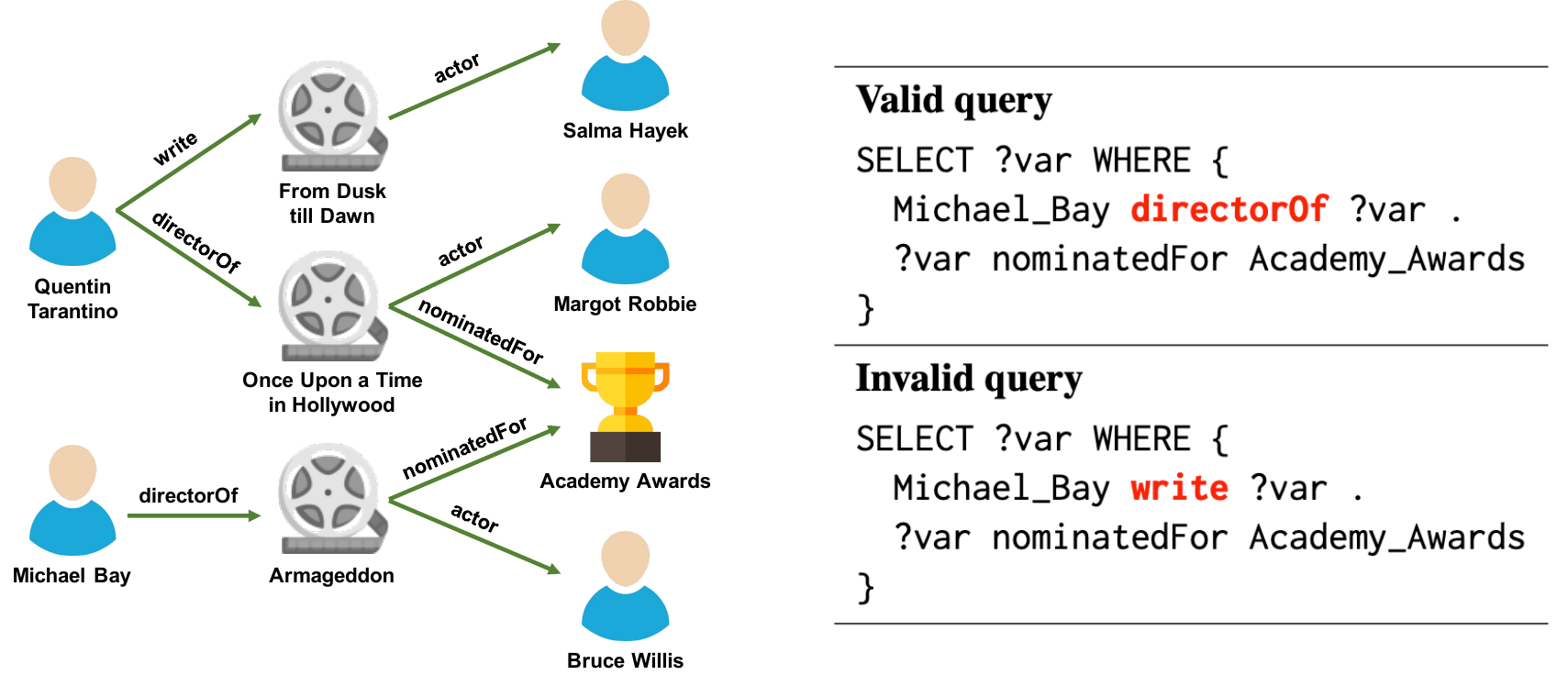}
    \caption{Valid and invalid SPARQL queries for the question \textbf{“What Michael Bay work has nominated for Academy Awards?"} within a knowledge graph (left). Since no movie is written by Michael Bay, it is invalid to generate (Michael\_Bay, write) pattern in a query.}
    \label{fig:example}
\end{figure*}

\section{Related Work}\label{related_work}

\subsection{Multi-Stage Approaches for KBQA}
In the field of Knowledge Base Question Answering (KBQA), numerous researchers have concentrated on multi-stage approaches.
These methods decompose the KBQA process into several stages, including entity linking, relation prediction, subgraph retrieval and query structure generation.
Such segmentation allows for handling the complexity of large-scale data sources more effectively.
PullNet \cite{sun-etal-2019-pullnet} iteratively identifies question-specific entities, constructs a subgraph, and then finds answers using a Graph Convolutional Network.
EDGQA \cite{edgqa2021Hu} decomposes the input question into an entity description graph using rules, from which subqueries are generated.
DECAF \cite{yu2023decaf} proposes text-based retrieval and generates logical form through Fusion-in-Decoder \cite{izacard-grave-2021-leveraging}.
Many studies utilize pre-trained language models (PLMs) for query structure generation \cite{das-etal-2021-case,hirigoyen-etal-2022-copy,modern,stagqa,ye-etal-2022-rng,chen2022outlining,DBLP:conf/esws/BanerjeeNUB23,qdtqa,NLQxform,KGQAn}, combined with independent entity linking modules.
\citet{li-etal-2023-shot} generates query drafts through Large Language Model (LLM) and refine them by linking entities with BM25 and FACC1 \cite{facc1}.
FlexKBQA \cite{li2023flexkbqa} also utilizes an LLM, but for generating synthetic data.
It then trains a teacher model, incorporating entity linking results from \citet{li-etal-2020-efficient}.

The performance of such multi-stage approaches is intrinsically tied to the outcomes of preceding stages.
Longer inference time also happens as a consequence.
The key distinction between multi-stage studies and SPARKLE is attributed to its end-to-end framework.
This ensures that the performance is influenced solely by a single model, thereby reducing inference times.

\subsection{End-to-End Approaches for KBQA}
End-to-end methods in KBQA leverage a single neural network model, providing an advantage in terms of simplicity.
Recent studies have incorporated PLMs to generate queries or retrieve direct answers.
Rigel-E2E \cite{saffari-etal-2021-end} jointly performs entity resolution and inference using a differentiable knowledge graph construction suggested by \citet{Cohen2020Scalable}.
KG-Flex \cite{McKenna2023} decodes into a  continuous embedding space where relations are expressed in natural language, enabling use of new relations at test time without retraining.
Similar to our approach, some studies \cite{soru-marx-nampi2018,sgpt} uses PLMs to generate full SPARQL queries with various linguistic features to overcome the complexity in multi-stage approaches.
However, these methods can generate invalid triple patterns and is inherently bound to the knowledge base used during training when conducting inference.
On the other hand, SPARKLE utilizes the structural information embedded in the knowledge base in real-time, enabling adaptive inference on the evolving knowledge base.

\subsection{Semantic Parsing with Constraints}

Semantic parsing methods often integrates constraints with a language model to ensure the output is grammatically correct.
PICARD \cite{scholak-etal-2021-picard} provides multiple levels of constraints, from lexical to grammatical, for text-to-SQL generation.
Pangu \cite{gu-etal-2023-dont}, on the other hand, utilizes an external symbolic agent that extends S-expressions iteratively.
TIARA \cite{shu-etal-2022-tiara} focuses on constrained decoding to generate valid KB classes and relations.
ArcaneQA \cite{gu-su-2022-arcaneqa} employs constrained decoding to narrow down the search space following pre-defined expansion rules.

Nonetheless, these KBQA approaches \cite{gu-etal-2023-dont,gu-su-2022-arcaneqa} that apply constraints on S-expressions still encounter difficulties in converting S-expressions to SPARQL queries \cite{hu-etal-2022-logical}.
SPARKLE focuses on generating a valid triple pattern by directly integrating knowledge base in the model without complex pre-defined rules or external components.
This method ensures SPARKLE naturally captures semantic structure of knowledge base during ongoing left-to-right query generation since SPARQL arranges a triple pattern in the order of $(h,r,t)$: "$h$ performs $r$ on $t$".

\section{Preliminaries}

A Knowledge Graph (KG), denoted as $\mathcal{G}$ is a collection of triples where each triple consists of three elements: a subject entity denoted as \emph{h}, a relation denoted as \emph{r}, and an object entity denoted as \emph{t}.
Formally, it is represented as $\{(h,r,t) | h, t \in \mathcal{E}, r \in \mathcal{R}\}$, where $\mathcal{E}$ represents the set of entities and $\mathcal{R}$ denotes the set of relations.
Each triple signifies the existence of a relational connection between the head entity \emph{h} and the tail entity \emph{t}.

The term SPARQL is an acronym for \emph{SPARQL Protocol and RDF Query Language}.
This query language plays an important role in a knowledge graph systems, enabling users to retrieve and manipulate data stored in RDF (Resource Description Framework) format.
SPARQL consists of query forms (e.g., ASK, SELECT, CONSTRUCT, DESCRIBE), modifiers (e.g., ORDER, PROJECTION, DISTINCT etc), and triple patterns.

It is important to note that the validity of a SPARQL query hinges on the existence of the specified triple patterns in the underlying RDF data.
For instance, consider a user asking, "What Michael Bay work has nominated for Academy Awards?" in a knowledge graph like Figure \ref{fig:example}.
A typical sequence-to-sequence model for SPARQL generation may include a triple pattern (Michael\_Bay, write, ?var) in a query, interpreting `write' as synonymous with `make'.
However, this pattern would be invalid because Michael Bay has no record of writing any movies in the underlying graph.
This example highlights the significance of incorporating real linkage information from the knowledge graph when generating SPARQL queries.
\begin{figure*}[t]
    \centering
    \subfloat[Generative entity relation retrieval]{\includegraphics[width=.480\textwidth, height=1.8cm]{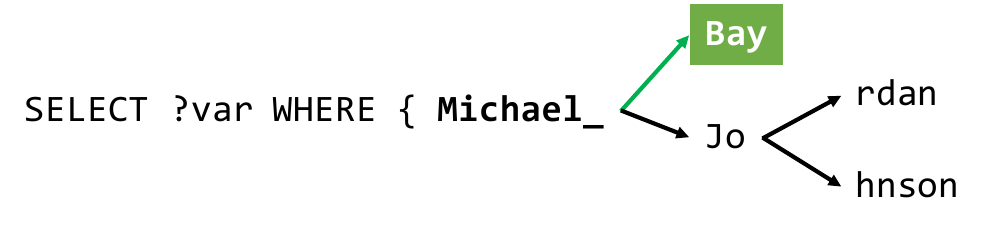}
      \label{fig:method_trie}
    }\hfill
    \subfloat[Pruning invalid triple patterns]{\includegraphics[width=.498\textwidth, height=1.8cm]{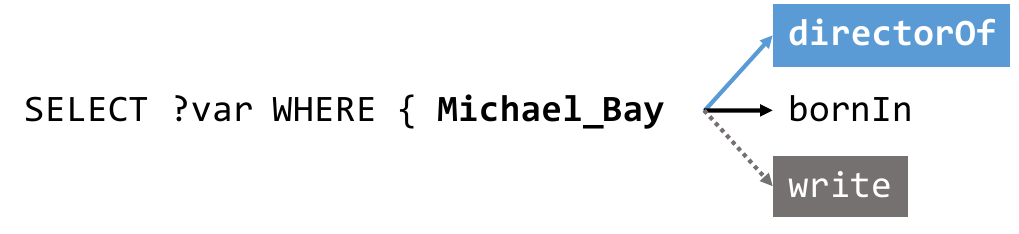}
      \label{fig:method_link}
    }
    \caption{Illustrative example of constrained decoding for SPARQL query generation using the input "\textbf{What Michael Bay work has nominated for Academy Awards?}". (a) Entity and relation retrieval: Our model generates entities and relations from the entity prefix trie and relation prefix trie. (b) Leveraging knowledge graph structure: During the generation of triple patterns, our model exploits the linkage information present in the KG. For instance, if the model is generating a relation for the head entity \textit{Michael\_Bay}, it excludes relation \textit{write} from candidates since there is no association of \textit{write} with \textit{Michael\_Bay} in the knowledge graph, as depicted in Figure \ref{fig:example}.}
    \label{fig:total_overview}
\end{figure*}

\section{Method}

We address the natural language to SPARQL problem using a single sequence-to-sequence model with constrained decoding during inference.
As depicted in Figure \ref{fig:total_overview}, our decoding involves two types of constraints.
We first focus on generation of entity and relation.
Additionally, we leverage the interconnected structure of the knowledge base when decoding relations after the head entity and, correspondingly, tail entities after the relation in triple patterns.
As highlighted by \citet{genre}, making entity and relation identifiers meaningful is crucial for the success of a sequence-to-sequence model to retrieve them.
Considering that knowledge bases such as Wikidata and Freebase use arbitrary identifiers, we convert these into more intuitive, human-readable identifiers using their names and types. 
For instance, we transform Quentin Tarantino's Freebase ID, m.0693l, into a human-readable format "\texttt{[quentin tarantino (film director)]}" (details in Appendix \ref{sec:identifier}).
SPARKLE is trained using a standard sequence-to-sequence objective, which aims to maximize the likelihood of the output sequence.

\subsection{Entity and Relation Generation}

For the generation of entity and relation, SPARKLE employs an autoregressive formulation that assigns a score to each entity $e \in \mathcal{E}$ and relation $r \in \mathcal{R}$, denoted as $p(z|x)$.

\fontsize{10}{9.5}{
\begin{equation}
    p(z|x) = \prod\limits_{i=1}^{n} p_{\theta}(y_i|x,y_1,...,y_{i-1})
\end{equation}
}

Here, $y$ represents the set of tokens in the identifiers of $e \in \mathcal{E}$ and $r \in \mathcal{R}$, $x$ is the input, $\theta$ stands for the model parameters, and $z$ denotes either $e$ or $r$.
To navigate the search space effectively, we make use of Beam Search \cite{beam-search} decoding strategies.
Currently, many multi-stage KBQA methods \cite{ye-etal-2022-rng,hu-etal-2022-logical,stagqa} exploit a dense retriever such as BLINK \cite{wu-etal-2020-scalable} and ELQ \cite{li-etal-2020-efficient}.
This approach, however, leads to retrieval costs increasing linearly with the growth of the knowledge base, as each input sentence must be compared against all entities or relations.
Instead, we simply rank multiple SPARQL queries that contain various entities and relations using Beam Search.
This significantly reduces the cost associated with the retrieval.
The time required for this process is now dependent on the size of beams and the length of identifiers, making it more manageable.

To enforce SPARKLE to generate only valid identifiers for entities and relations, we define identifier tries $ \mathcal{T} $ as described by \citet{genre}.
An example of such a trie is depicted in Figure \ref{fig:method_trie}.
Each node within $ \mathcal{T} $ is annotated with tokens from the vocabulary.
As the model traverses the nodes in the trie starting from the root, it generates a next token based on the previous ones.
Therefore, each child node represents all the possible continuous tokens required to construct valid identifiers.

\subsection{Pruning Invalid Triple Patterns}

We additionally extend constrained decoding to prevent the model from generating invalid triple patterns within SPARQL queries.
There are two scenarios where we can exploit the structural information of the knowledge base.
The first case occurs when generating a relation after the head entity has been generated.
In this case, we constrain the decoding process to consider only relations that are linked to the head entity.
The probability formulation for this scenario is as follows:
\fontsize{10.0}{9.5}{
\begin{equation}
    p(r \in \mathcal{N}(e^{head}) | x) = \prod\limits_{i=1}^{n} p_{\theta}(w_i|x,w_{<s},e^{head}_{s,\dots,i})
\end{equation}
}
where $w$ denotes the set of tokens available, $e^{head}$ represents the head entity, and $\mathcal{N}$ indicates the set of neighbors.
The subscript $s...i$ is used to denote the indices of the most recently generated identifier tokens because each identifier in the output of language model is represented by a series of tokens rather than a single token.
For instance, if the model has generated "\texttt{SELECT ?var \{ [ Michael\_Jordan ]}" up to this point, it's understood that the head entity identifier comprises multiple tokens (e.g., \texttt{[, Michael\_, Jo, rdan, ]} ).
In this notation, $s$ is used to mark the index of the first token of the head entity, enabling the model to accurately identify the entity within the ongoing query generation. 

The second case concerns the generation of a tail entity after the relation has been generated.
Here, we restrict the tail entity decoding process to consider only entities that have a connection to the specified relation in the knowledge base.
The probability formulation for this scenario is as follows:
\fontsize{10.0}{9.5}{
\begin{equation}
    p(e^{tail} \in \mathcal{N}(r) | x) = \prod\limits_{i=1}^{n} p_{\theta}(w_i|x,w_{<s}, r_{s,\dots,i})
\end{equation}
}

The applicability and effectiveness of these strategies are rooted in the nature of SPARQL, which organizes triple patterns in the order of $(h,r,t)$.
This structure is inherently compatible with the left-to-right decoding of sequence-to-sequence models, making it feasible to implement these constraints efficiently.
We enforce these constraints by masking the log probabilities of tokens (setting their score to $-\inf$) for invalid entities, relations, and connectivities.
Our approach not only ensures the syntactical accuracy of the generated triple pattern but also aligns them with the semantic structure of the underlying knowledge base.

\section{Experiments}

\subsection{Experimental Setting}

\subsubsection{Dataset}
In our experiments, SPARKLE is evaluated across three benchmark datasets, each sourced from different knowledge bases: LCQuAD 1.0 \cite{lcquad1}, WebQSP \cite{webqsp} and SimpleQuestions-Wiki \cite{simplequestions}.
SimpleQuestions-Wiki, a large-scale KBQA dataset, provides 14,184 train questions, 2,111 dev questions and 4,116 test questions, annotated on Wikidata \cite{wikidata}.
We use Wikidata dump from December 2017 and filter out questions whose triples are not supported by the dump.
LCQuAD 1.0 comprises 5,000 questions accompanied by corresponding SPARQL queries, each of which can be answered using DBpedia \cite{dbpedia} 2016-04.
We allocate 200 questions from the 4,000 in the training dataset to a dev dataset.
WebQSP consists of 4,937 questions designed for semantic parsing on Freebase \cite{freebase}.
We adopt the same train and dev splits employed by \citet{yu2023decaf}.

\subsubsection{Evaluation Metrics}
Following previous works \cite{stagqa,ye-etal-2022-rng,yu2023decaf}, our model is evaluated using Hits@1 and F1 score.
Both metrics function as a comprehensive gauge of our model's ability to retrieve answer sets in KBQA.

\subsubsection{Implementation Details}
We first build identifier tries using Marisa trie, a memory-efficient trie, designed by \citet{yata2011dictionary}.
This is important because a knowledge base can have millions of components.
To make sure that looking up connections in the knowledge base does not slow down the decoding, we store this information in a hash table.
To ensure that the model can effectively handle SPARQL expressions, we tokenize each SPARQL terms (e.g., `?var', `SELECT' and `ORDER BY') as individual tokens.
When generating entities, we insert variables to the candidates so that the model can choose to generate them.
For training, we fine-tune the pre-trained BART model \cite{lewis-etal-2020-bart} with a sequence-to-sequence objective, maximizing $ log p_{\theta}(y|x) $ with respect to model’s parameters $ \theta $, which is commonly used in neural machine translation.
In the inference phase, we use beam search to generate the top-k SPARQL queries.
These queries are executed in turn until a non-empty query result is obtained, following the previous works \cite{hu-etal-2022-logical,shu-etal-2022-tiara}.
For more training details, refer to Appendix \ref{sec:experimental_details}.

\subsection{Overall Performance}
\label{sec:overall}

{
\newcolumntype{C}[1]{>{\centering\arraybackslash}m{#1}}
\begin{table*}[t]
  \small
  \centering
  \begin{tabular}{l c c}
    \toprule
     & \textbf{SimpleQuestions \ -Wiki} & \textbf{LCQuAD 1.0} \tabularnewline
    \midrule
     \textbf{Method} & \textbf{F1} & \textbf{F1} \tabularnewline
    \midrule
    Falcon 2.0 \cite{saker2020falcon}  & 36.3 & - \tabularnewline
    SYGMA \cite{neelam-etal-2022-sygma}  & 44.0 & - \tabularnewline
    KGQAN \cite{KGQAn} & - & 51.6 \tabularnewline
    EDGQA \cite{edgqa2021Hu}  & - & 53.1 \tabularnewline
    GETT-QA \cite{DBLP:conf/esws/BanerjeeNUB23} & 76.1 & - \tabularnewline
    STaG-QA \cite{stagqa}  & 61.2 & 53.6 \tabularnewline
    QDTQA \cite{qdtqa} & - & 58.8 \tabularnewline
    AQGNet* \cite{aqgnet} & - & 74.8 \tabularnewline
    HGNet* \cite{chen2022outlining} & - & \textbf{78.1} \tabularnewline
    \midrule
    \textbf{SPARKLE} & \textbf{79.6} & 72.2 \tabularnewline
    \hspace{5mm} w/o pruning & 78.7 & 64.9 \tabularnewline
    \hspace{5mm} w/o constraints & 65.5 & 57.3 \tabularnewline
    \bottomrule
  \end{tabular}
  \caption{F1 results on the test splits of benchmark datasets: SimpleQuestions-Wiki, LCQuAD 1.0. \textbf{Bold} indicates best model. * denotes using gold entity mentions.}
  \label{tab:overall}
\end{table*}
}

{
\newcolumntype{C}[1]{>{\centering\arraybackslash}m{#1}}
\begin{table}[h]
  \small
  \centering
  \begin{tabular}{p{5cm}C{0.7cm}C{0.7cm}}
    \toprule
     \textbf{Method} & \textbf{Hits@1} & \textbf{F1} \tabularnewline
    \midrule
    \multicolumn{3}{c}{\textit{multi-stage, LLM}} \tabularnewline
    PullNet \cite{sun-etal-2019-pullnet} & 68.1 & - \tabularnewline
    STaG-QA \cite{stagqa} & 68.5 & - \tabularnewline
    HGNet* \cite{chen2022outlining} & 71.7 & - \tabularnewline
    RnG-KBQA \cite{ye-etal-2022-rng} & - & 75.6 \tabularnewline
    DECAF \cite{yu2023decaf} & \textbf{80.7} & \textbf{77.1} \tabularnewline
    StructGPT* \cite{jiang-etal-2023-structgpt} & 72.6 & - \tabularnewline
    FC-KBQA \cite{zhang-etal-2023-fc} & - & 76.9 \tabularnewline
    \midrule
    \multicolumn{3}{c}{\textit{end-to-end}} \tabularnewline
    Rigel-E2E \cite{saffari-etal-2021-end} & 45.0 & - \tabularnewline
    ReifKB \cite{Cohen2020Scalable} & 52.7 & - \tabularnewline
    KG-Flex \cite{McKenna2023} & 68.9 & - \tabularnewline
    \midrule
    \textbf{SPARKLE} & 71.2 & 71.1 \tabularnewline
    \hspace{5mm} w/o pruning & 70.3 & 70.1 \tabularnewline
    \hspace{5mm} w/o constraints & 64.0 & 63.9 \tabularnewline
    \bottomrule
  \end{tabular}
  \caption{F1 and Hits@1 results on WebQSP. \textbf{Bold} indicates best model. * denotes using gold entity mentions.}
  \label{tab:overall_webqsp}
\end{table}
}

We evaluate SPARKLE with a variety of KBQA methods, including but not limited to natural language to SPARQL models.
In Table \ref{tab:overall}, we compare SPARKLE with multi-stage methods on two benchmark datasets\footnote{To our knowledge, no previous work has applied an end-to-end natural language to SPARQL method to SimpleQuestions-Wiki and LCQuAD 1.0.}.
Our findings highlight that SPARKLE excels on SimpleQuestions-Wiki and LCQuAD 1.0, setting a new state-of-the-art on SimpleQuestions-Wiki.
SPARKLE exhibits competitive results on LCQuAD 1.0, aligning closely with the prior SOTA models.
It is noteworthy that unlike HGNet \cite{chen2022outlining} and AQGNet \cite{aqgnet}, which assume gold entities are given, our model works without this feature.
Therefore, SPARKLE is positioned as a leading model among those not utilizing gold entities.

In Table \ref{tab:overall_webqsp}, we evaluate SPARKLE against both multi-stage and end-to-end methds on WebQSP.
The performance of SPARKLE on WebQSP is not impressive as other two datasets.
This stems from the prevalence of complexity in the SPARQL queries in WebQSP.
The dataset contains intricate expressions like those related to time (dates or periods), sequences (ordering elements), and string comparisons.
These types of queries often require multiple advanced SPARQL syntax elements such as type casting, ORDER BY and FILTER.
For these complex queries, SPARKLE scores an F1 of 0.38, which is noticeably lower than its performance on simpler questions. 
Given that SPARKLE primarily focuses on constrained decoding of triple patterns, the model faces challenges in generating theses complex conditions within a simple sequence-to-sequence framework.
Nevertheless, it is noteworthy that SPARKLE achieves the highest Hits@1 score among end-to-end methods, demonstrating its efficiency despite its simplicity.

\begin{figure*}
    \centering
    \includegraphics[width=0.9\linewidth]{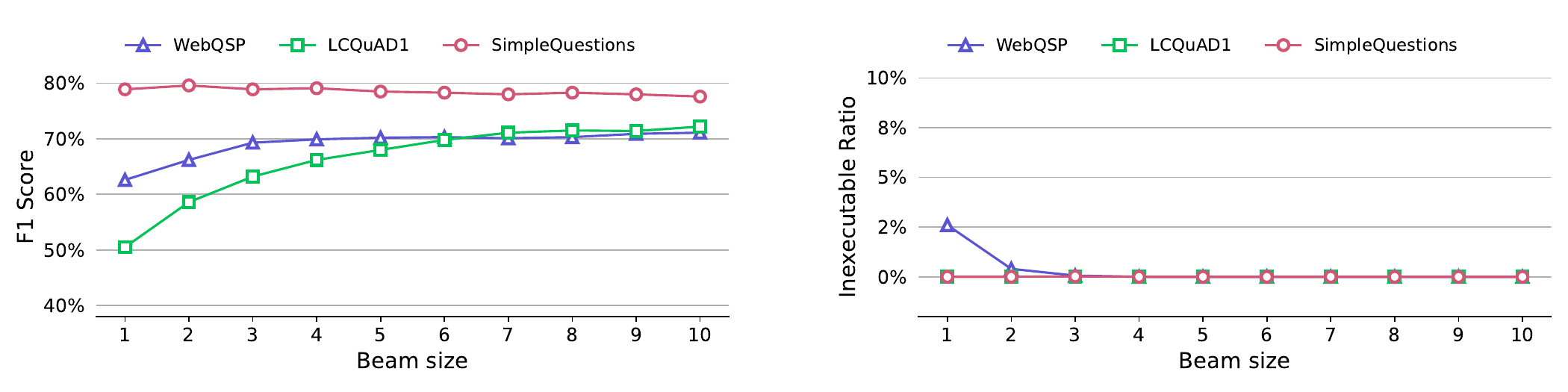}
    \caption{The influence of beam size on model performance and the proportion of inexecutable queries generated on benchmark datasets.}
    \label{fig:beam_size}
\end{figure*}

\subsection{Analysis}

\subsubsection{Impact of Constrained Decoding}

To assess the effect of the proposed constrained decoding, we conduct ablation studies in Table \ref{tab:overall}, \ref{tab:overall_webqsp}.
SPARKLE without pruning refers to our model only applying constraint on retrieving entities and relations in a generative way, without utilizing structural information for pruning.
SPARKLE without constraints describes our model operating entirely unconstrained, identical to a BART model during inference.
The results clearly demonstrate a gradual decrease in performance when constraints are removed in sequence.
Specifically, the absence of any constraints leads to a significant performance decline, with a reduction of up to 14.9 F1 scores observed in the LCQuAD 1.0 dataset.

Moreover, there is a noticeable increase in the proportion of queries that cannot be executed.
Without any constraints, the ratio of inexecutable queries rises to 27.9\%, 12.4\%, and 17.5\% for SimpleQuestions-Wiki, LCQuAD 1.0, and WebQSP, respectively.
This indicates that a standalone sequence-to-sequence model struggles to accurately generate valid identifiers of entity and relation, as well as triple patterns.
Therefore it becomes clear that constrained decoding is crucial for the generation of valid query components.

\subsubsection{Impact of Beam Size}

In Figure \ref{fig:beam_size}, we present a comprehensive overview of how beam size influences the performance of our model across three benchmark datasets.
On LCQuAD 1.0 and WebQSP, we observe a significant improvement in the model's performance as beam size increases.
This improvement appears to reach a saturation point when the beam size approaches 7.
Beyond this point, further increases in beam size yield diminishing returns in terms of performance enhancement.

When the beam size is set to 1, representing a greedy decoding, the model is prone to making incorrect predictions for entities and relations.
This problem arises due to the abundance of entities that share similar label and type in the knowledge base, leading to similar identifiers.
Such similarities pose challenges for the model in accurately retrieving the correct entity.
This initial retrieval failure has a cascading effect on the model's ability to prune invalid triple patterns since subsequent retrievals are based on the preceding, potentially erroneous choices.
Consequently, increasing beam size acts as a safeguard against such erroneous selections, ultimately reducing the likelihood of generating inaccurate triple patterns within the query.

Nonetheless, increasing the beam size does not yield similar improvements on SimpleQuestions-Wiki.
SPARQL queries in SimpleQuestions-Wiki involve only a single triple pattern, and many questions explicitly include the surface forms of entity and relation.
It seems increasing beam size rather degrades the model performance \cite{pmlr-v97-cohen19a} on such straightforward questions.

Another noteworthy observation is that the rate of inexecutable queries remains consistently near zero, irrespective of the beam size, indicating that simple constrained decoding is effective in ensuring the generation of executable queries.

\begin{table*}[t]
    \begin{tcolorbox}
    \centering
    \begin{tabularx}{\textwidth}{ X }
        \addlinespace[0.1cm]
        \textbf{Case I Question:} "What is the result of AlphaGo versus Lee Sedol?" \tabularnewline
        
        \small\textbf{Prediction (w/ DBPedia 16-10):} \texttt{SELECT DISTINCT ?uri WHERE \{ \textcolor{red}{\textbf{<http://dbpedia.org/resource/AlphaGo\_\break versus\_Lee\_Sedol>}} <http://dbpedia.org/property/result> ?uri \}} \tabularnewline
        
        \addlinespace[0.2cm]
        
        \textbf{Case II Question:} "How many awards have Bob Dylan got?" \tabularnewline
        
        \small\textbf{Prediction (w/ DBPedia 16-10):} \texttt{SELECT DISTINCT COUNT(?uri) WHERE \{ <http://dbpedia.org/resource/\break Bob\_Dylan> \textcolor{red}{\textbf{<http://dbpedia.org/ontology/award>}} ?uri \}} \tabularnewline

    \end{tabularx}
    \end{tcolorbox}
    \caption{Case study of predictied SPARQL query on events regeistered on DBPedia between April 2016 to October 2016.}
    \label{tab:cold-start}
\end{table*}

{
\newcolumntype{C}[1]{>{\centering\arraybackslash}m{#1}}
\begin{table}
    \small
    \centering
    \begin{tabular}{l c c c}
        \toprule
         \textbf{Dataset} & \textbf{Seen} & \textbf{Unseen} & \textbf{Total}\\
        \midrule
        \textbf{SimpleQ} & 90.3 \tiny(138) & 79.2 \tiny(3973) & 79.6 \tiny(4111) \\
        \textbf{LCQuAD} & 78.3 \tiny(423) & 67.7 \tiny(577) & 72.2 \tiny(1000) \\
        \textbf{WebQSP} & 84.9 \tiny(939) & 51.5 \tiny(658) & 71.1 \tiny(1597) \\
        \midrule
        \textbf{Total} & 83.5 \tiny(1500) & 74.4 \tiny(5208) & 76.4 \tiny(6708) \\
        \bottomrule
    \end{tabular}
    \caption{Evaluation of SPARKLE on seen and partially/entirely unseen data across test set.}
    \label{tab:dataset-cold-start}
\end{table}
}

\subsubsection{Adaptive Inference}

SPARKLE, in its operation, dynamically uses the structure of knowledge base at runtime.
This straightforward approach empowers SPARKLE to make adaptive inferences based on a knowledge base that differs from the one used during its training.
We initially trained SPARKLE using LCQuAD 1.0 with DBPedia 2016-04 dump. 
Subsequently, we put SPARKLE to the test by performing inferences using DBPedia 2016-10 dump, representing a knowledge base that evolved over time.
To assess the model's adaptability, we manually create two questions that are related to the events occurred between April 2016 and October 2016.
These questions include newly introduced entity and newly linked relation.
The results of theses evaluations are presented in Table \ref{tab:cold-start}.

In Case I, SPARKLE successfully retrieves the newly registered entity \texttt{\seqsplit{<http://dbpedia.org/resource/AlphaGo\_versus\_Lee\_Sedol>}} which was not a part of the knowledge base during the model's training.
In Case II, when SPARKLE performs inference using DBPedia 2016-04, it is unable to generate a pattern involving \texttt{\seqsplit{<http://dbpedia.org/resource/Bob\_Dylan>}} and \texttt{\seqsplit{<http://dbpedia.org/ontology/award>}}.
This is due to the fact that the relation \texttt{\seqsplit{<http://dbpedia.org/ontology/award>}} was added to the entity \texttt{\seqsplit{<http://dbpedia.org/resource/Bob\_Dylan>}} after he received the Nobel Prize in Literature in October 2016.
These examples vividly illustrate SPARKLE's ability to perform adaptive inference when confronted with newly introduced entities and relations without retraining.

In Table \ref{tab:dataset-cold-start}, we additionally evaluate SPARKLE's ability to adapt with queries involving partially or entirely unseen entities or relations.
Each test set is split into two categories: Seen and Unseen.
Unseen refers to queries where at least one entity or relation is not encountered during training.
Despite new entities and relations, SPARKLE shows robust performance on such questions.

\subsubsection{Inference time}

We assess the efficiency of SPARKLE by measuring the average inference time per question on the test split of each dataset.
The distinctive feature of SPARKLE lies in its single sequence-to-sequence model architecture, which results in fast inference speed as shown in Table \ref{tab:inference_time}.
The experiments used an NVIDIA V100 GPU and a beam size of 10 during decoding.

Since many KBQA approaches employ multi-stage methods, they often suffer from longer inference time \cite{gu-etal-2021-bertranking,ye-etal-2022-rng,shu-etal-2022-tiara}.
Multi-stage approaches involve the loading and unloading of data to and from the GPU when processing questions in a sequential manner.
A direct comparison of inference times between SPARKLE and other KBQA models is challenging for several reasons.
Different models often employ varying datasets and some send queries via SPARQL endpoint (e.g., Virtuoso server) for intermediate computations.
Given these complexities, our analysis focuses on the marginal increase in latency by SPARKLE over a naive sequence-to-sequence model.
This slight increase stems from the computational overhead of masking probabilities for invalid tokens during auto-regressive generation.
Nonetheless, employing a Trie structure for retrieving identifiers and a hash table for managing connectivity significantly mitigates complexity.
Moreover, as discussed in Section \ref{sec:overall}, the model's performance stabilizes with a beam size of 7, allowing faster inference without significant performance loss.

Additionally, one of the key strengths of SPARKLE is its ability to perform batch processing.
As an end-to-end system, it can handle multiple queries simultaneously, potentially improving throughput in practical applications.
As shown in Table \ref{tab:inference_time}, employing batch processing with a batch size of 8 reduces the average inference time per query by up to 41\%.
This feature is especially beneficial in real-world scenarios where handling large volumes of queries is essential.

{
\newcolumntype{C}[1]{>{\centering\arraybackslash}m{#1}}
\begin{table}
    \small
    \centering
    \begin{tabular}{l c c c}
        \toprule
         \textbf{Dataset} & \textbf{w/o CD} & \textbf{SPARKLE} & \textbf{Batch (8)} \\
        \midrule
        \textbf{SimpleQ} & 0.30 & 0.37 & 1.74 ($41\% \uparrow$) \\
        \textbf{LCQuAD} & 0.45 & 0.63 & 3.16 ($37\% \uparrow$) \\
        \textbf{WebQSP} & 0.45 & 0.94 & 5.83 ($22\% \uparrow$) \\
        \bottomrule
    \end{tabular}
    \caption{Average inference time per query (seconds) and speed increase with batch processing (percentage).}
    \label{tab:inference_time}
\end{table}
}
\section{Conclusion}

In this work, we present SPARKLE, a novel end-to-end approach that directly use the structural information of knowledge base to enhance SPARQL query generation.
SPARKLE employs a straightforward yet effective strategy of constrained decoding in two contexts: retrieving entities and relations in a generative way, and pruning invalid triple patterns based on knowledge base structure.
The experimental results show that our approach helps sequence-to-sequence models generate executable queries, resulting in strong performance across benchmark datasets: SimpleQuestions-Wiki, LCQuAD 1.0 and WebQSP.
Moreover, SPARKLE's adaptability is demonstrated as it can accommodate new entities and relations without retraining, simply by switching the underlying knowledge base during inference.
We additionally show that SPARKLE offers faster inference time and supports batch processing, allowing simultaneous handling of multiple questions.

\section*{Limitations}

Although our approach shows good performance with its simple architecture, there remains scope for further enhancements.
Our model requires substantial memory resources to utilize structural information of knowledge base during decoding.
As the size of the knowledge base grows, these memory requirements increase linearly.
Considering that large knowledge bases often contain over millions of entities, managing the connectivity information for such knowledge base becomes challenging.
However, for large platforms providing KBQA services, the bigger challenge is not memory resources; rather, it is delivering a real-time service to their users.

Moreover, our use of constrained decoding is currently restricted to the generation of triple patterns.
While triple pattern is the most important component of SPARQL queries, SPARQL itself comprises more advanced expressions.
For instance, a bottom-up constrained parsing for nested queries, such as those involving UNION clause, could improve our model.
A comprehensive SPARQL query generation requires addressing these additional components.
Looking ahead, we plan to enhance our models by incorporating such grammatical analysis of SPARQL to support more sophisticated query constructs.

\bibliography{anthology,custom}

\appendix

\section{Human-readable identifiers}
\label{sec:identifier}

Table \ref{tab:identifier} provides examples of how we convert entity IRIs into human-readable identifiers.
While DBPedia \cite{dbpedia} includes the entity's label and type in its IRIs, rendering them already meaningful, Wikidata \cite{wikidata} and Freebase \cite{freebase} use random characters for their IRIs.
To make these IRIs more interpretable, we extract the label and type of an entity from the respective knowledge bases and format the identifiers as \texttt{label(type)}.
For Wikidata entities, we determine their type using the \texttt{P31(instance of)} relation.
In Freebase, \texttt{common.topic.notable.types} is used for this purpose.
When an entity is associated with multiple types, we randomly select two of these for construction of the identifier.
Additionally, if an entity shares its label and type with others, we append its IRI to the end of the identifier to ensure uniqueness.
To clearly differentiate entities and relations from other text elements, we enclose the identifier with square brackets at its beginning and end.

\setlength{\belowcaptionskip}{10pt}

{
\newcolumntype{C}[1]{>{\centering\arraybackslash}m{#1}}
\begin{table*}
    \centering
    \begin{tabular}{C{2cm}|C{3.5cm}|C{5.5cm}}
        \hline
         \textbf{KB} & \textbf{Entity IRI} & \textbf{Human-readable Identifier} \\
        \hline
        DBPedia & Quentin\_Tarantino & [ quentin tarantino : resource ] \\
        Wikidata & Q3772 & [ quentin tarantino (human) ] \\
        Freebase & m.0693l & [ quentin tarantino (film director) ] \\
        \hline
    \end{tabular}
    \caption{Examples of entity IRI and human-readable identifiers in SPARKLE.}
    \label{tab:identifier}
\end{table*}
}

{
\newcolumntype{C}[1]{>{\centering\arraybackslash}m{#1}}
\begin{table*}
    \centering
    \begin{tabular}{l | c | c | c}
        \hline
          & \textbf{SimpleQuestions} & \textbf{LCQuAD 1.0} & \textbf{WebQSP} \\
        \hline
        \textbf{Training time (h)} & 24.1 & 6.8 & 8.0 \\
        \textbf{Emission (kgCO2eq)} & 3.8 & 1.1 & 1.3 \\
        \textbf{Memory usage(GB)} & 54.5 & 18.7 & 147.1 \\
        \hline
    \end{tabular}
    \caption{Details on training costs.}
    \label{tab:training_cost}
\end{table*}
}

\section{Training Details}
\label{sec:experimental_details}

SPARKLE is developed using PyTorch \cite{pytorch} and HuggingFace library \cite{wolf2019huggingface} for both training and inference.
Throughout the training process, Adam optimizer \cite{adam} is utilized.
The model's learning rate is determined through experiments and searched from [5e-4, 5e-5, 5e-6].
Our training objective is a sequence-to-sequence cross-entropy loss without label smoothing.
The models are trained using a batch size of 32.
The experiments are conducted on 4 to 8 NVIDIA V100 GPUs.
For the output generation, the maximum token length is set to 128 for all datasets.
Table \ref{tab:training_cost} shows training and evaluation costs of SPARKLE in the perspective of time, space and emission.

\newcommand{\spaces}[1]{\foreach \\ in {1,...,#1}{\ }}

\spaces{5} \\
\spaces{5} \\
\spaces{5} \\
\spaces{5} \\
\spaces{5} \\
\spaces{5} \\
\spaces{5} \\
\spaces{5} \\
\spaces{5} \\
\spaces{5} \\
\spaces{5} \\
\spaces{5} \\
\spaces{5} \\
\spaces{5} \\
\spaces{5} \\
\spaces{5} \\
\spaces{5} \\
\spaces{5} \\
\spaces{5} \\
\spaces{5} \\
\spaces{5} \\
\spaces{5} \\
\spaces{5} \\
\spaces{5} \\
\spaces{5} \\
\spaces{5} \\
\spaces{5} \\
\spaces{5} \\
\spaces{5} \\
\spaces{5} \\
\spaces{5} \\

\end{document}